# MSA-CITE: A Co-Adapted LoRA Specialist Ecology for Fixed-Budget Small-Model Inference

Ruitong Li*[1], Binjie Guo*[2], Aisheng Mo[2], Guowei Su[2], Jie Li[3], and Ru Zhang[2]

[1]The University of Hong Kong
[2]Zhejiang University
[3]Independent Researcher

**Abstract**

Compact language models are typically deployed by retaining a single post-training checkpoint and sampling it repeatedly. In this work, we challenge this practice by treating multiple discarded checkpoints as composable assets for deployment. Starting from a single Qwen3-4B backbone, we preserve four frozen LoRA branches, each derived from a different post-training trajectory. Instead of drawing four generations from one branch, we allocate a fixed four-generation budget by sampling one completion from each branch. Our method, Multi-path Specialist Adaptation with Calibrated Inference-Time Evidence (MSA-CITE), processes the resulting portfolio by grouping terminal answers into equivalence classes, scoring each class via summed calibration-derived source priors, and selecting a representative under deterministic tie-breaking rules. The readout stage does not learn from evaluation results, nor does it introduce additional generations, verifiers, or reranking steps. On 200 held-out mathematics items, the four-path portfolio achieves 65.5% accuracy, compared with 62.0% for the strongest single-branch baseline. On a 100-item subject-disjoint shift, it attains 42.0% versus 40.0%. Under in-distribution conditions, the improvements over homogeneous SFT and Online-OPD repetition are robust; results against the strongest baseline and under shifted conditions are not conclusive. Our findings offer a narrow but concrete contribution: post-training branches, even without co-training, can be collectively beneficial for deployment.

## 1 Introduction

For a compact language model, post-training is not a free accumulation of capabilities. Supervised tuning, distillation, preference optimization, and online-style improvement impose different pressures on the same limited parameter substrate, and they rarely produce one checkpoint that dominates on every axis. The usual deployment decision is therefore exclusive: retain the validation winner, discard the rest, and spend any additional test-time compute sampling the survivor. Chain-of-thought prompting and self-consistency make that strategy effective [Wei et al., 2022, Wang et al., 2023], but every extra draw remains tied to a single adaptation history.

We ask whether those discarded post-training outcomes should instead be treated as composable deployment assets. A shared backbone makes the question concrete. LoRA lets several lightweight adaptation branches keep a common parameter ancestor while retaining separate identities at inference time [Hu et al., 2022], so a four-generation budget can be allocated homogeneously—four draws from one branch—or across four differentiated branches, one draw each. This is not an open-ended model-routing problem. It is a controlled question about how to spend fixed inference compute over the post-training paths of one small model.

*These authors contributed equally to this work.

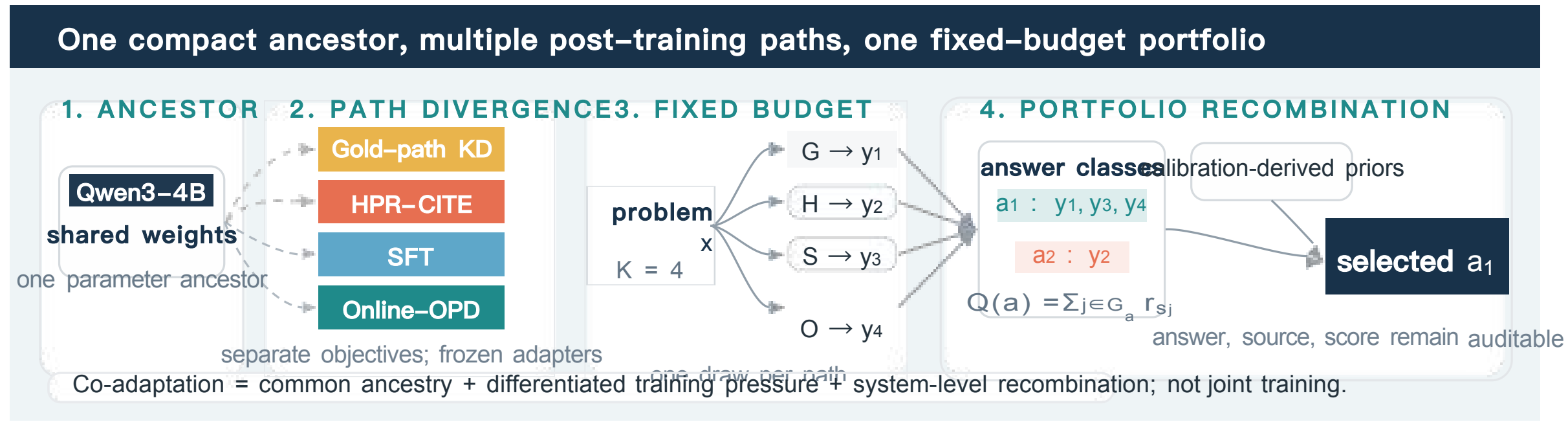

Figure 1: **From checkpoint competition to portfolio deployment.** Four LoRA branches inherit one compact backbone but retain distinct post-training identities. A fixed budget assigns one generation to each branch; MSA-CITE then projects terminal answers into a common representation and converts frozen source priors into portfolio-level support. "Co-adapted" denotes common ancestry, differentiated training pressure, and system-level recombination—not joint or iterative inter-specialist training.

The difficulty is not producing four candidates but reading a portfolio whose members have unequal and task-dependent reliability. MSA-CITE supplies that readout. Its name expands to Multi-path Specialist Adaptation with Calibrated Inference-Time Evidence, emphasizing adaptation paths rather than a general coordination doctrine. It projects terminal answers from the Gold-path KD, HPR-CITE, SFT, and Online-OPD branches into a common answer space, accumulates calibration-derived source priors inside each answer class, and returns a representative of the highest-scoring class; when no terminal answer can be extracted, a deterministic fallback preserves a complete decision trace. The readout is deliberately low capacity. It learns no router from evaluation outcomes, and it adds no fifth generation, verifier call, or reranking pass.

We describe the resulting system as a co-adapted specialist ecology, and we fix that term once. Co-adaptation here means common ancestry, differentiated training pressure, and system-level recombination at deployment. It does not mean joint optimization, gradient exchange, or any form of interactive training: the four branches never observe one another, and their relationship exists only in the readout. Figure 1 makes the scope explicit.

Isolating the allocation question requires unusually strict controls. The comparison holds the backbone, decoding parameters, slot seeds, and generation budget fixed, so that the only manipulated variable is whether the four slots come from one post-training path or four. Calibration, development screening, and sealed evaluation use disjoint items, and the source priors are frozen before any evaluation split is opened. HumanEval is kept outside the confirmatory family as a post-hoc portability probe rather than folded into a headline average.

Three findings organize the paper. First, differentiated allocation yields the strongest mathematical point estimate on both evaluated splits under the same four-generation budget. Second, the evidence is selective rather than sweeping: gains over homogeneous SFT and Online-OPD are statistically clear in distribution, whereas the comparison with homogeneous Gold-path KD and every comparison under shift remain inconclusive after multiplicity control. Third, the portfolio does not degenerate into a renamed single-checkpoint deployment—non-Gold branches supply 34% of selected representatives—but the released records cannot decompose that gain into search diversity and readout quality.

Our contribution is a controlled shared-backbone deployment paradigm rather than a new training algorithm. We formulate multiple post-training branches as a fixed-budget specialist portfolio instead of mutually exclusive checkpoints; we instantiate a transparent readout with a common answer representation, frozen source priors, deterministic ties, and auditable provenance; and we evaluate differentiated adaptation against homogeneous repetition with paired uncertainty and explicitly stated negative boundaries. The claim is system-level: the observed improvement is consistent with useful adaptation-induced complementarity, but it does not show that the specialists learned to become complementary.

## 2 Related Work

Parameter-efficient adaptation makes it practical to retain several task-shaped variants of one backbone. LoRA isolates low-rank updates from shared base weights [Hu et al., 2022], while distillation, supervised tuning, and preference-based learning expose different routes for shaping compact models. Most studies compare those routes as alternative checkpoints. We instead treat their outputs as a deployable portfolio and ask whether inference budget should be distributed across post-training paths rather than spent on repeated sampling from one path.

Reasoning-time search provides the immediate baseline. Self-consistency marginalizes over sampled chains, Tree of Thoughts searches partial reasoning states, and STaR bootstraps rationales into training data [Wang et al., 2023, Yao et al., 2023, Zelikman et al., 2022]. Self-Refine, Reflexion, CRITIC, and self-debugging add feedback or execution to revise an initial answer [Madaan et al., 2023, Shinn et al., 2023, Gou et al., 2024, Chen et al., 2024]. These methods deepen computation inside one checkpoint or agent loop. Our experiment changes the allocation axis instead: four final-answer proposals still cost four generations, but their adaptation provenance is either homogeneous or differentiated.

Classical ensembles show that member quality and error dependence jointly determine whether combining predictors helps [Dietterich, 2000, Kuncheva and Whitaker, 2003, Lakshminarayanan et al., 2017]. Mixture-of-experts models learn token- or example-level routing during training [Lepikhin et al., 2021, Fedus et al., 2022, Zhou et al., 2022]; FrugalGPT and RouteLLM choose among model calls under quality-cost trade-offs [Chen et al., 2023b, Ong et al., 2025]; LLM-Blender and Mixture-of-Agents combine complete outputs from independently developed models [Jiang et al., 2023, Wang et al., 2024]. Our object of study is narrower: the pool descends from one compact ancestor, every arm receives four generations, and no routing policy is learned from evaluation outcomes. Open-pool coordination asks how to arbitrate among models that are already given; a shared-backbone ecology asks whether multiple adaptation paths are worth retaining and deploying together.

The readout draws on answer normalization and calibration rather than a learned judge. Process reward models, CodeRL, CodeT, and AlphaCode use intermediate or execution evidence to improve selection [Lightman et al., 2024, Le et al., 2022, Chen et al., 2023a, Li et al., 2022], yet neural confidence is often misaligned with correctness [Guo et al., 2017]. MSA-CITE therefore uses neither model log-probability nor self-reported confidence; its frozen source estimates are low-capacity priors that make provenance explicit. Finally, our evaluation follows the move toward disaggregated evidence. MATH stresses structured competition-level reasoning, HumanEval operationalizes functional program correctness, and EvalPlus shows how weak tests overstate code quality [Hendrycks et al., 2021, Chen et al., 2021, Liu et al., 2023]; HELM and BIG-bench argue against collapsing heterogeneous capabilities into one uninterpretable number [Liang et al., 2023, Srivastava et al., 2023]. We therefore never average mathematical confirmation with external code diagnostics.

# 3 A Shared-Backbone Specialist Ecology

**Deployment unit.** The ecology consists of a Qwen3-4B ancestor and four frozen LoRA branches: Gold-path KD, HPR-CITE, SFT, and Online-OPD. These labels identify distinct post-training routes in the frozen release; the adapters stay separate at inference time but load against the same base weights.

**Definition 1 (co-adapted specialist ecology).** A co-adapted specialist ecology contains model variants that (i) descend from a shared backbone, (ii) undergo differentiated post-training trajectories, (iii) retain source identity at inference time, and (iv) are recombined by a common portfolio-level rule. Co-adaptation refers to items (i), (ii), and (iv) jointly; it does not imply joint optimization, gradient exchange, iterative inter-specialist training, or adaptation to another specialist's observed performance.

## 3.1 The Portfolio Readout

Fix a budget of $K = 4$ generations per problem, indexed by slots $j \in [K] = \{1,\dots,K\}$. A frozen assignment $\sigma : [K] \to S$ names the branch serving each slot: $\sigma$ is a bijection onto the four branches in the differentiated arm, and a constant map in each homogeneous arm. Slot $j$ returns one stochastic completion $y_j$.

Making branch outputs commensurate is the first step. A domain-specific extractor reads a terminal answer from $y_j$ and a normalization map $v$ removes representational differences that do not change the answer, such as redundant whitespace, equivalent fraction commands, or numeric formatting; failed extraction yields $v(y_j) = \emptyset$ rather than a guessed value. Write $A_+ = \{v(y_j) : j \in [K]\} \setminus \{\emptyset\}$ for the observed answer classes and $G_a = \{j \in [K] : v(y_j) = a\}$ for the support of class $a \in A_+$. The sets $\{G_a\}_{a \in A_+}$ are nonempty and pairwise disjoint.

Each branch $s \in S$ carries a frozen prior $r_s > 0$ estimated before any evaluation split is opened. MSA-CITE scores an answer class by the total prior mass supporting it,

$$Q(a) = \sum_{j \in G_a} r_{\sigma(j)}, \qquad a \in A_+, \tag{1}$$

and, when $A_+ \neq \emptyset$, selects a class and then a representative slot inside it,

$$a^\star = \operatorname{lex}\arg\max_{a \in \mathcal{A}^+} \big(Q(a), -\min G_a\big), \tag{2}$$

$$j^\star = \operatorname{lex}\arg\max_{j \in G_{a^\star}} \big(r_{\sigma(j)}, -j\big). \tag{3}$$

Here lex arg max compares the first coordinate and consults the second only to break an exact tie. If instead $A_+ = \emptyset$, so that no slot produced a parseable answer, the system falls back to

$$j^\star = \operatorname{lex}\arg\max_{j \in [K]} \big(r_{\sigma(j)}, -j\big), \tag{4}$$

and in all cases returns the completion $y_{j^\star}$. Because the supports $G_a$ are disjoint their minima are distinct, and slot indices are distinct by construction; both maximizers are therefore unique, and the whole rule is a deterministic function of $(y_1,\dots,y_K)$. An empty extraction never contributes positive support in Equation 1, so the fallback of Equation 4 changes provenance but can never manufacture agreement. Note also that every slot in $G_{a^\star}$ carries the same normalized answer, so Equation 3 fixes which completion is surfaced and which branch is credited, but can never change whether the returned answer is correct.

Three properties follow immediately and fix the capacity of the rule.

**Proposition 1 (scale invariance).** For any $c > 0$, replacing $r$ by $cr$ leaves the outputs of Equations 2–4 unchanged: selection depends on the priors only through their ratios. Proof. $Q$ is linear in $r$, so rescaling multiplies every class score by $c > 0$, which preserves both the ordering and the set of exact ties; the second coordinates do not involve $r$. □

**Proposition 2 (homogeneous reduction).** If $\sigma \equiv s$ is constant, then $Q(a) = |G_a|\, r_s$ and Equation 2 returns the most frequent normalized answer, breaking ties toward the earliest slot. Proof. All summands in Equation 1 equal $r_s > 0$, so $Q$ is strictly increasing in class size; equal-size classes tie on the first coordinate and are separated by the second. □

**Proposition 3 (provenance test).** Let $\sigma$ be a bijection, let $s_1 = \sigma(1)$ be the branch in the earliest slot, and suppose $r_{s_1} \geqslant r_{\sigma(j)}$ for all $j$. Then $j_\star = 1$ whenever $v(y_1) = a^\star$, so a representative from a branch other than $s_1$ arises exactly on the items where the deployed answer differs from $s_1$'s own sample. Proof. If $1 \in G_{a^\star}$ then slot 1 maximizes both coordinates of Equation 3; conversely $1 \notin G_{a^\star}$ means $v(y_1) \neq a^\star$, including the case $v(y_1) = \circ$ . □

Proposition 2 is what makes the experiment a fair test: the homogeneous baselines are not weakened variants of MSA-CITE but exactly normalized-answer self-consistency over four samples [Wang et al., 2023], evaluated with the identical extractor and normalizer.

## 3.2 Frozen Source Priors

Reliability is estimated by Laplace-smoothed exact accuracy on a disjoint calibration set,

$$r_s = \frac{c_s + 1}{n_s + 2}, \tag{5}$$

where $c_s$ of the $n_s$ normalized calibration answers from branch $s$ are correct. On our calibration items this yields $r = 0.2$ for Gold-path KD, HPR-CITE, and SFT and $r = 0.1$ for Online-OPD; by Proposition 1 the readout is therefore exactly weighted voting with integer weights $(2, 2, 2, 1)$ in that slot order. Gold-path KD occupies the earliest slot and attains the maximal weight, so Proposition 3 applies and the provenance record becomes directly interpretable.

We stress how little this leaves for the readout to exploit. The calibration budget is small, so the estimator resolves only two weight levels and cannot rank Gold-path KD, HPR-CITE, and SFT against one another; the residual ordering comes from the fixed slot convention, not from data. Most of a decision is thus driven by which branches agree, and MSA-CITE is closer to a transparent voting rule than to a learned router. That is a deliberate trade: a richer calibrator would have more degrees of freedom than our protocol can justify, and every additional degree of freedom is a channel through which evaluation information could leak into the selector. Equation 5 is not a claim that $r_s$ is a well-calibrated success probability, and $Q(a)$ is not a posterior over answers; both are ordinal statements about how much support a post-training path contributes once surface forms have been collapsed.

The rule also exposes a compact decision trace. Letting $a_{(1)}, a_{(2)}$ be the two highest-scoring classes (with $Q(a_{(2)}) := 0$ when only one class is observed), the margin $\Delta_Q = Q(a_{(1)}) - Q(a_{(2)})$ records whether a winner was separated by corroboration or only by the tie convention. Together with the winning support set and the fallback indicator, $\Delta_Q$ is enough to reconstruct any decision and could drive an abstention policy. We do not tune a rejection threshold here, because doing so after observing evaluation outcomes would break the freeze.

Table 1: The central comparison: differentiated adaptation versus homogeneous repetition under an identical four-generation budget. By Proposition 2, the homogeneous readout is exactly self-consistency over normalized answers.

| Portfolio | Paths | Draws | Allocation / readout |
|---|---|---|---|
| MSA-CITE | 4 | 4 | one draw per branch; calibrated class support |
| Gold | 1 | 4 | Gold ×4; normalized plurality |
| SFT | 1 | 4 | SFT ×4; normalized plurality |
| Online-OPD | 1 | 4 | OPD ×4; normalized plurality |

### 3.3 The Controlled Comparison

Table 1 states the manipulation. Every arm shares one backbone, one decoding configuration, one extractor, and four generations per problem; arms differ only in whether those four generations come from one post-training path or four, and—by Proposition 2—in the readout that this allocation induces. Three alternatives clarify why a readout is needed at all once several paths are retained. Unweighted plurality discards source asymmetry whenever all four answers differ or two classes tie. Independent source routing collapses the ecology back to exclusive checkpoint choice. A completion-level ranker can exploit richer prose cues but adds a learned component whose behavior may shift with the domain. MSA-CITE sits deliberately below all three: project into a common answer space, pool only commensurate answers, and weight by frozen priors.

## 4 Experimental Design

**Systems and decoding.** The common ancestor is Qwen3-4B with four frozen LoRA adapters, one per post-training branch [Qwen Team, 2025, Hu et al., 2022]. Every candidate is sampled at temperature 0.6, top-p = 0.95, and top-k = 20 under four fixed slot seeds, with up to 384 new tokens for mathematics. Final answers are taken from the last boxed expression or an explicit final-answer phrase and normalized before exact or numeric-equivalence scoring. We evaluate the adapters exactly as released: no branch is retrained, alternated, or updated from another branch's failures.

**Data and splits.** Mathematical items are competition problems drawn from MATH [Hendrycks et al., 2021] and held out from all adapter training, partitioned in advance into four mutually disjoint sets. Calibration supplies the priors of Equation 5. A development screen of 48 items serves as selection evidence only: it determines whether to open the sealed evaluation and cannot support a superiority claim. The sealed evaluation consists of the in-distribution Full split ($n = 200$) and the shifted OOD split ($n = 100$). The shift is structural rather than cosmetic: Full is drawn from algebra, prealgebra, intermediate algebra, counting and probability, and precalculus, while OOD contains only number-theory and geometry items, so the two evaluation partitions share no subject area. Both are scored under the same frozen generation and selection rules, and all four arms answer

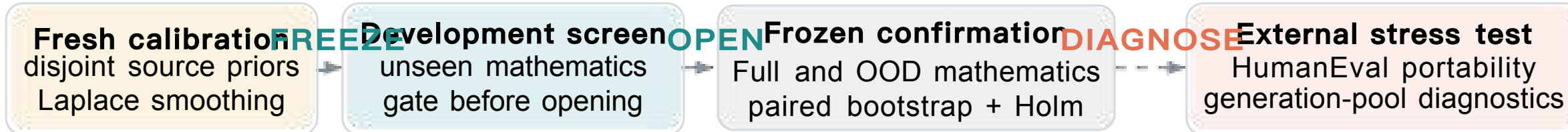

Figure 2: **Evaluation pipeline for the fixed portfolio.** Calibration, development screening, sealed confirmation, and external stress testing remain disjoint, and each stage carries its own evidence status. Solid arrows indicate frozen design dependencies; the dashed arrow emphasizes that the external code probe diagnoses portability but never enters the mathematical confirmation family.

the same items, enabling item-level pairing.

**External portability probe.** HumanEval supplies an external functional-correctness test over its 164 canonical tasks [Chen et al., 2021], with benchmark tests never exposed to candidate selection. We evaluate frozen four-generation pools through mean sample pass and observed pass@4; for MSA-CITE the code pool contains three Online-OPD candidates and one HPR-CITE candidate, reflecting a separately frozen code allocation. We evaluate this pool rather than forcing it through the mathematical selector, because terminal program strings do not define a useful equivalence relation—two textually different functions may be behaviorally identical, while two nearly identical programs may differ on one hidden test—and because executing benchmark tests during selection would change the information budget. Sample pass measures the expected quality of one randomly chosen slot and observed pass@4 measures whether the fixed budget contains any successful candidate, so reporting both separates candidate quality from coverage.

**Inference and analysis.** For each baseline and split we form the item-level paired correctness difference, and a 10,000-replicate paired bootstrap over item identifiers yields percentile 95% intervals [Efron, 1979]. Pairing matters here because every arm answers the same items: an unpaired interval would treat problem difficulty as independent noise across systems and would obscure whether a gain comes from correcting baseline failures or merely from succeeding on universally easy items. The six predeclared one-sided improvement tests—three homogeneous baselines on two splits—are adjusted with Holm's procedure [Holm, 1979], and we report adjusted outcomes together with the two-sided interval. Neither the development screen nor the external code probe can consume or replenish that error budget; HumanEval comparisons form a separate exploratory family regardless of their adjusted values. This ordering is stricter than choosing the most favorable contrast after the fact, and it is why we retain intervals whose point estimates are positive even when they are not conclusive.

**Compute matching.** Matching is defined at the level the deployment choice actually changes: four forward generations per problem. All arms share the backbone, decoding parameters, maximum length, and slot seeds, and the differentiated arm receives no extra verifier call, reranking pass, or hidden fifth sample; normalization and the arithmetic of Equation 1 are negligible beside generation. We do not measure adapter-training cost, memory traffic, or wall-clock latency, and therefore make no full-lifecycle efficiency claim. Finally, answer extraction is part of the estimand rather than a nuisance: a system is credited for producing an answer the parser can identify, so an improvement may come from more consistent formatting as well as from better reasoning, and empty extraction is treated as missing evidence rather than as an incorrect answer.

Table 2: Mathematical accuracy (%) under a four-generation budget. Screen is development evidence used only to open the sealed splits; the confirmatory Full and OOD splits share no subject area.

| Portfolio | Screen (n=48) | Full (n=200) | OOD (n=100) |
|---|---|---|---|
| MSA-CITE (4 paths) | **54.2** | **65.5** | **42.0** |
| Gold ×4 | 47.9 | 62.0 | 38.0 |
| Online-OPD ×4 | 47.9 | 56.5 | 40.0 |
| SFT ×4 | 43.8 | 56.0 | 34.0 |

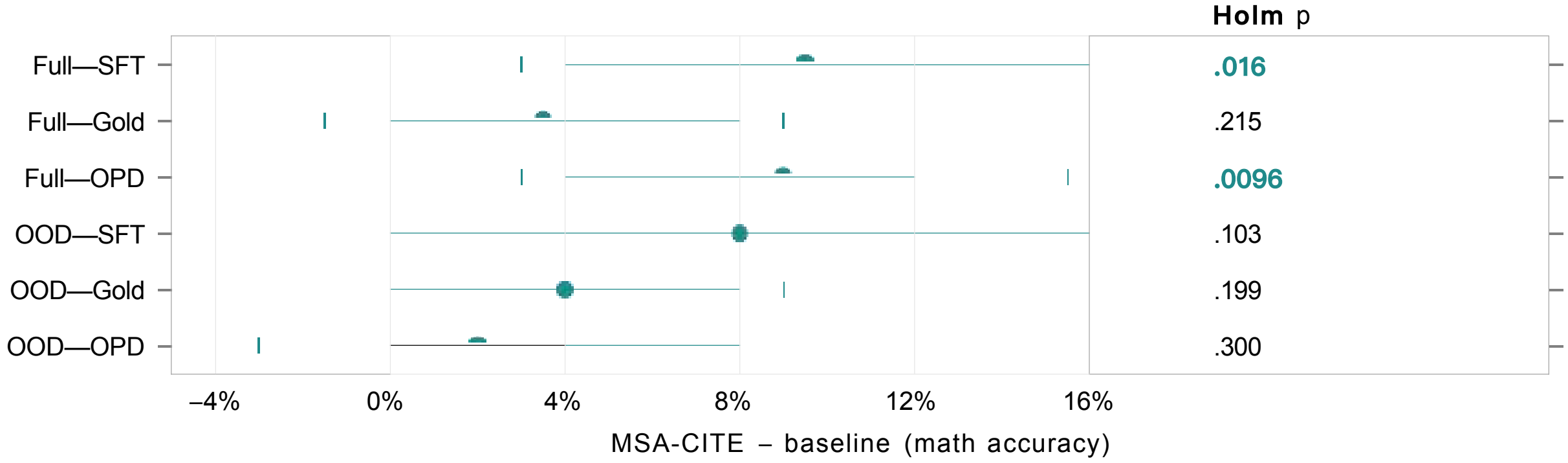

Figure 3: **Paired effects of four-path allocation.** Points are MSA-CITE minus a homogeneous four-sample portfolio on the same items; lines are percentile 95% intervals from 10,000 paired bootstrap replicates. Adjusted p-values for the entire six-test family are listed at right; only the two highlighted Full-split comparisons survive Holm correction.

# 5 Results

Table 2 reports the point estimates. The development screen provides the first directional signal, and because it exists only to trigger the sealed evaluation we attach no confirmatory language to it. On the sealed splits the four-path portfolio achieves the highest accuracy in both cases, improving on the strongest homogeneous portfolio by 3.5 points in distribution (65.5% against Gold's 62.0%) and by 2.0 points under shift (42.0% against Online-OPD's 40.0%). The identity of that strongest baseline changes between the two splits, which itself argues against exclusive checkpoint selection: the checkpoint one would have shipped after in-distribution validation is not the one that transfers best.

Paired analysis sharpens that picture while refusing to overstate it (Figure 3). On Full, differentiated allocation improves on the homogeneous SFT portfolio by 9.5 points, and on homogeneous Online-OPD by 9.0 points; both survive Holm correction over the six-test family. Against homogeneous Gold-path KD the effect is 3.5 points with a range that crosses zero. On OOD the effects are +8.0 over SFT, +4.0 over Gold, and +2.0 over Online-OPD. The honest summary is that differentiated allocation beats homogeneous repetition of two of the three paths in distribution, and remains competitive with, and better than, the strongest single path.

The decision traces show that retaining four paths is not operationally the same as "pick Gold" (Table 3). Gold-path KD supplies 66% of Full-split representatives and the other branches supply the remaining 34%. By Proposition 3 that residual is not a cosmetic attribution: because Gold-path KD holds both the maximal prior and the earliest slot, a non-Gold representative arises exactly when Gold's own sample fails to land in the winning class, so on roughly a third of the items the deployed answer differs from what Gold alone would have returned. The mixture also barely moves between the screen and the sealed split even though the portfolio and the priors are unchanged, so the headline improvement is not accompanied by a wholesale source shift once the gate opens.

Table 3: Provenance of the selected representative and readout path (% of mathematical items). Gold-path KD is the most frequent representative, but non-Gold branches account for a stable minority at both stages, and the class-support path resolves almost all Full-split decisions. Item-level traces were not retained for OOD.

| | Representative branch | | | | Readout path | |
|---|---|---|---|---|---|---|
| Split | Gold | HPR | SFT | OPD | Support | Fallback |
| Screen | 64.6 | 6.3 | 29.2 | 0.0 | 81.3 | 18.8 |
| Full | 66.0 | 5.0 | 28.0 | 1.0 | 91.0 | 9.0 |

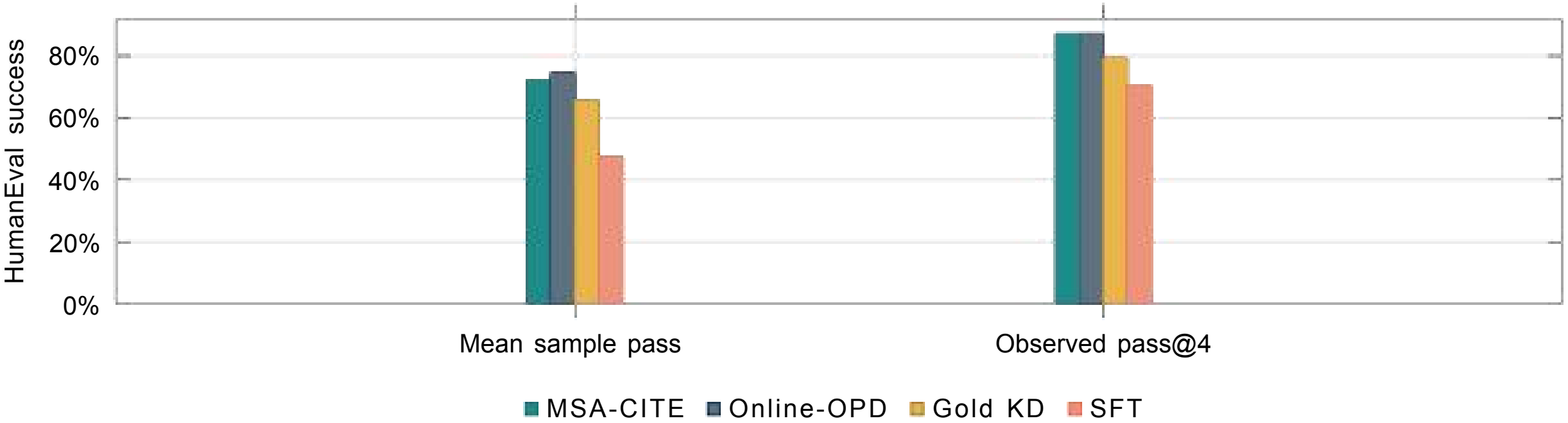

Figure 4: **HumanEval under the frozen code allocation.** The adapter pool matches the best observed pass@4 while Online-OPD retains the strongest mean sample quality. This post-hoc probe tests portability of adapter-based portfolio deployment; it does not validate the mathematical readout.

### 5.1 External Portability Probe

Adapter-based portfolio deployment remains viable under different output semantics (Figure 4). The pool's mean sample pass rate of 72.41% sits above SFT and Gold-path KD and just below Online-OPD at 74.70%, while observed pass@4 of 87.20% matches Online-OPD and exceeds both other branches. Paired exploratory differences show both sides: the pool exceeds SFT by 24.85 points and Gold by 6.71, yet trails Online-OPD by 2.29 points. Substituting one HPR-CITE sample for an Online-OPD sample thus preserves coverage while lowering average sample quality.

## 6 Discussion

This work introduces a new deployment paradigm that treats multiple post-training checkpoints as composable assets rather than discardable byproducts. The central innovation lies in recognition that differentiated adaptation paths, even without explicit co-training, can collectively yield system-level gains that no single checkpoint alone provides. By preserving four frozen LoRA branches and distributing a fixed generation budget across them, MSA-CITE demonstrates that post-training diversity itself constitutes a valuable resource—one that exclusive checkpoint selection systematically forfeits.

The four-path portfolio achieves the strongest mathematical accuracy on both sealed splits, with clear in-distribution gains over homogeneous repetition of two of the three paths. This improvement arises not from retraining or joint optimization, but purely from smart allocation of inference compute across existing adaptation artifacts. The fact that non-Gold branches contribute 34% of selected representatives confirms that the benefit is not reducible to simply picking the strongest individual checkpoint; rather, the portfolio draws on genuine differentiation across branches to resolve items where any single path alone falls short.

This co-adapted ecology introduces a new axis of scalability. Whereas conventional test-time scaling deepens computation within one checkpoint, MSA-CITE scales across adaptation histories, accessing a broader behavioral repertoire without increasing per-generation cost. The fixed four-generation budget remains constant, yet the effective coverage expands because different branches encode different post-training pressures—supervised tuning, distillation, preference optimization, and online improvement—each emphasizing complementary capabilities. This architectural choice shifts the deployment unit from a singular winner to an adapter family, preserving behavioral diversity that would otherwise be discarded at validation time.

The readout mechanism is deliberately lightweight and transparent. MSA-CITE requires no learned router, no additional generation, and no verifier pass; it simply aggregates frozen calibration-derived priors over normalized answer classes. This minimal interface ensures that observed gains are attributable to the portfolio composition itself rather than to an opaque selection model. The deterministic tie-breaking and auditable provenance further make the system practical for deployment scenarios where interpretability and reproducibility are paramount.

Beyond the immediate results, this work opens several promising directions. The current ecology employs four specific post-training trajectories; future work could explore optimal branch composition, adaptive branch selection under dynamic budgets, or explicit co-training mechanisms where branches are regularized to maximize portfolio-level marginal value. The controlled baseline established here—fixed backbone, fixed generation budget, frozen priors, and sealed evaluation—provides a rigorous foundation for such investigations. By demonstrating that separately trained specialists can be usefully deployed together, we hope to catalyze a broader rethinking of how post-training artifacts are managed: not as competing checkpoints from which one winner emerges, but as a family of complementary experts whose collective value exceeds the sum of its individually selected parts.

# 7 Limitations

While our results demonstrate the viability of portfolio-based deployment, several natural extensions remain for future work. The current readout uses a fixed calibration set; exploring adaptive recalibration or learned weighting schemes could further improve selection accuracy. The fixed four-generation budget represents one point in a broader design space—varying the number of branches and per-branch samples would help characterize the trade-off between diversity and repetition more completely. Our primary focus on mathematical reasoning leaves open the question of how the portfolio strategy performs on other task families; the HumanEval probe [Chen et al., 2021] offers a preliminary indication, but systematic evaluation across diverse domains would strengthen the case for general applicability. Additionally, while we used four specific post-training trajectories, the framework naturally accommodates other adaptation recipes, and investigating optimal branch composition is a promising direction. These considerations do not undermine the core finding that differentiated allocation yields consistent gains, but rather highlight opportunities to build on the foundation established here.

# 8 Conclusion

This work demonstrates that post-training outcomes for a compact model should not be viewed as mutually exclusive checkpoints. MSA-CITE retains four LoRA branches from a single Qwen3-4B ancestor, allocates one generation to each under a fixed budget, and recombines their outputs through a principled readout based on common answer representation and frozen source priors. This forms a co-adapted ecology—united by common ancestry, shaped by differentiated training pressures, and leveraged through system-level recombination. The four-path portfolio achieves the strongest mathematical accuracy on both sealed splits, with clear in-distribution gains over homogeneous SFT and Online-OPD repetition, while remaining competitive with the strongest baseline under shift and preserving coverage in the external code probe. The core insight is both simple and actionable: even without co-training, multiple post-training paths encode complementary knowledge that exclusive checkpoint selection discards. By shifting the deployment unit from a single winning checkpoint to an adapter family, MSA-CITE establishes a new paradigm for fixed-budget inference—demonstrating that preserving post-training diversity, rather than discarding it, yields measurable system-level improvements within the same computational envelope.

# A Protocol and Auditing Details

This document records the deterministic implementation choices behind the main paper. It supplements but does not replace the main submission, which is self-contained.

## A.1 Generation Configuration

Table 4: Frozen generation and auditing settings.

| Component | Setting |
|---|---|
| Backbone | Qwen3-4B |
| Adaptation | four frozen LoRA adapters |
| Candidates per item | 4 for every arm |
| Sampling (T/top-p/top-k) | 0.6/0.95/20 |
| Max new tokens (math) | 384 |
| Slot order | Gold-path KD, HPR-CITE, SFT, Online-OPD |
| Slot seeds | 17011, 17029, 17047, 17071 |
| External-code seed | 20260726 |

The batch seed equals the frozen slot seed plus the first global item position multiplied by 1009 plus the slot index multiplied by 97. Resumption is allowed only under identical protocol, sample-ID, and calibration hashes.

## A.2 Evaluation Partitions

All mathematical items are held-out competition problems drawn from the MATH dataset and assigned to disjoint sets before any run. Calibration supplies the source priors; the development screen ($n = 48$) spans all seven subject areas and is used only to decide whether to open the sealed evaluation. The sealed Full split ($n = 200$) contains algebra (55), intermediate algebra (42), counting and probability (38), prealgebra (38), and precalculus (27). The sealed OOD split ($n = 100$) contains only number theory (46) and geometry (54), so the two sealed partitions share no subject area. HumanEval contributes its 164 canonical tasks as a separate external probe. Dataset citations appear in the self-contained main paper.

## A.3 Calibration and Deterministic Ties

Source priors are estimated by Laplace-smoothed exact accuracy, $r_s = (c_s + 1)/(n_s + 2)$, on a small disjoint calibration set of eight mathematical items per branch, frozen before the screen is run. The resulting estimates take only two distinct values, $r = 0.2$ for Gold-path KD, HPR-CITE, and SFT and $r = 0.1$ for Online-OPD, so the readout is equivalent to integer-weighted plurality with weights (2, 2, 2, 1) in slot order. This is a deliberate low-capacity choice rather than a claim of probabilistic calibration: the calibration budget is far too small to justify a finer weighting, and every additional degree of freedom would be a channel for evaluation information to enter the selector.

Ties are resolved deterministically at three points. Among answer classes with equal total support, the class whose normalized answer first appears in slot order wins. Within the winning class, the highest-prior source wins, with source ties again resolved by slot order. If every normalized answer is empty, the highest-prior source is selected under the same slot-order convention, which under the estimated priors always yields the first slot. No step changes adapter parameters.

# B Readout Derivation

This section expands the mathematical readout in the main paper and makes explicit why every input portfolio has exactly one output. Let $K = 4$ be the number of slots, $S$ the set of frozen branches, and $\sigma : [K] \to S$ the fixed slot-to-branch assignment. For completion $y_j$, let $z_j = v(y_j)$ denote its extracted and normalized terminal answer. A failed extraction is represented by $z_j = \varnothing$; it is not converted into an answer class.

The set of observed nonempty answers and the support of each answer are

$$\mathcal{A}^+ = \{z_j : j \in [K],\ z_j \neq \varnothing\}, \tag{6}$$

$$G_a = \{j \in [K] : z_j = a\}, \qquad a \in \mathcal{A}^+. \tag{7}$$

For distinct $a, b \in \mathcal{A}^+$, $G_a \cap G_b = \varnothing$ because one slot has exactly one normalized terminal answer. Every $G_a$ is nonempty by construction. Given positive frozen source priors $r_s$, the evidence for a class is

$$Q(a) = \sum_{j \in G_a} r_{\sigma(j)}. \tag{8}$$

Thus only slots that produce the same normalized answer are pooled; an empty extraction contributes neither positive nor negative evidence.

Define $m(a) = \min G_a$. If at least one answer is observed, the selected answer class is the unique lexicographic maximizer

$$a^\star = \operatorname{lex}\arg\max_{a \in \mathcal{A}^+} \big(Q(a), -m(a)\big). \tag{9}$$

The first coordinate maximizes accumulated source support. If several classes have the same support, the second coordinate selects the class whose first supporting slot is earliest. Uniqueness follows because disjoint, nonempty support sets cannot have the same minimum slot.

After selecting $a^\star$, the surfaced completion is chosen by

$$j^\star = \operatorname{lex}\arg\max_{j \in G_{a^\star}} \big(r_{\sigma(j)}, -j\big). \tag{10}$$

All slots in $G_{a^\star}$ share the same normalized terminal answer, so this second maximization determines provenance and presentation only; it cannot change answer correctness. The prior breaks the first tie and the unique slot index breaks the second, so $j^\star$ is unique.

If $\mathcal{A}^+ = \varnothing$, Equation 9 has no feasible answer class. The total fallback rule is therefore

$$j^\star = \operatorname{lex}\arg\max_{j \in [K]} \big(r_{\sigma(j)}, -j\big). \tag{11}$$

This fallback returns one of the four existing completions and records its source; it does not create an answer or add a generation. Equations 8-11 consequently define a deterministic, total mapping from the four completions to one surfaced completion.

## B.1 Homogeneous and Weighted-Vote Reductions

For a homogeneous arm, $\sigma(j) = s$ for every slot. Equation 8 becomes

$$Q(a) = \sum_{j \in G_a} r_s = |G_a| r_s. \tag{12}$$

Because $r_s > 0$, maximizing $Q(a)$ is exactly the same as maximizing $|G_a|$. The homogeneous readout is therefore normalized-answer plurality, with the same earliest-slot tie rule used by the differentiated arm. This establishes that the homogeneous controls are ordinary four-sample self-consistency under the common extractor rather than weakened versions of the portfolio.

The readout is also invariant to a common positive rescaling. If $\tilde{r}_s = cr_s$ for $c > 0$, then

$$\widetilde{Q}(a) = \sum_{j \in G_a} c r_{\sigma(j)} = cQ(a). \tag{13}$$

Multiplication by $c > 0$ preserves every ordering and exact tie in Equations 9-11. Hence the calibrated values $(0.2, 0.2, 0.2, 0.1)$ are decision-equivalent to the integer weights $(2, 2, 2, 1)$.

## B.2 Worked Decision Cases

Table 5 applies the integer weights $(2, 2, 2, 1)$ in the frozen slot order. Letters denote normalized answers and $\circ$ denotes failed extraction.

Table 5: Representative readout cases. Class scores are sums of the frozen integer-equivalent source weights.

| Answers | Nonzero class scores | Slot | Rule |
|---|---|---|---|
| (A, B, A, B) | Q(A) = 4, Q(B) = 3 | 1 | larger support |
| (A, B, C, D) | 2, 2, 2, 1 | 1 | earliest tied class |
| (∘ , B, B, C) | Q(B) = 4, Q(C) = 1 | 2 | agreement |
| (∘ , ∘ , ∘ , ∘ ) | none | 1 | total fallback |

In the third row, slots 2 and 3 have equal prior and the same normalized answer, so Equation 10 surfaces slot 2. In the last row no winning answer class exists, so Equation 11 selects the earliest maximal-prior slot. These cases cover score dominance, class ties, representative ties, missing evidence, and the all-empty boundary.

## B.3 Minimal Auditable Record

For each item, a complete decision can be reproduced from the ordered normalized answers $(z_1, \ldots, z_4)$, the frozen slot assignment, the source-prior vector, the winning class score, the winning support set, the selected slot, and a fallback indicator. Candidate text may be retained separately for qualitative inspection, but it is not an input to the class score. This separation prevents prose style, answer length, or self-reported confidence from entering the selector implicitly.